\documentclass{article}

\usepackage{times}
\usepackage{graphicx} 

\usepackage{natbib}

\usepackage{algorithm}
\usepackage{algorithmic}

\usepackage{amsmath}
\usepackage{mathrsfs}
\usepackage{dsfont}
\usepackage{amssymb}
\usepackage{url}
\usepackage{caption}
\usepackage{multirow} 
\usepackage{amsthm}
\usepackage{thmtools} 
\usepackage{subcaption}
\usepackage{paralist}
\usepackage{fancyvrb}
 
\declaretheoremstyle[%
  headfont=\normalfont\itshape,%
  postheadspace=1em,%
  qed=\qedsymbol%
]{mystyle} 
\declaretheorem[name={Proof},style=mystyle,unnumbered,
]{spacelessprf}
\usepackage{setspace}

\usepackage[american]{babel}

\usepackage{makecell}
\usepackage{array}
\newcolumntype{M}[1]{>{\centering\arraybackslash}m{#1}}
\newcolumntype{L}[1]{>{\arraybackslash}m{#1}}

\newcommand {\w} {\boldsymbol{w}}
\newcommand {\W} {\boldsymbol{W}}
\newcommand {\y} {\boldsymbol{y}}
\newcommand {\Y} {\boldsymbol{Y}}
\newcommand {\x} {\boldsymbol{x}}
\newcommand {\bc} {\boldsymbol{c}}
\newcommand {\X} {\boldsymbol{X}}

\newcommand {\F} {\boldsymbol{F}}
\DeclareMathOperator*{\argmin}{arg\,min}
\DeclareMathOperator*{\argmax}{arg\,max}
\newcommand {\llbracket} {[\![}
\newcommand {\rrbracket} {]\!]}
\newtheorem {mydef} {Definition}
\makeatletter
\def\thm@space@setup{%
  \thm@preskip=0.2cm
  \thm@postskip=\thm@preskip 
}
\makeatother
\newtheorem {mythm} {Theorem}

\usepackage{relsize}

\renewcommand{\algorithmicrequire}{\textbf{Input:}}
\renewcommand{\algorithmicensure}{\textbf{Output:}}

\usepackage{pifont}
\newcommand{\xmark}{\text{\ding{55}}} 

\usepackage{hyperref}

\usepackage[accepted]{icml2017}

\icmltitlerunning{A Unified View of Multi-Label Performance Measures}

\begin{document} 

\twocolumn[
\icmltitle{A Unified View of Multi-Label Performance Measures}




\begin{icmlauthorlist}
\icmlauthor{Xi-Zhu Wu}{to}
\icmlauthor{Zhi-Hua Zhou}{to}
\end{icmlauthorlist}

\icmlaffiliation{to}{National Key Laboratory for Novel Software Technology, Nanjing University, Nanjing 210023, China}
\icmlcorrespondingauthor{Zhi-Hua Zhou}{zhouzh@lamda.nju.edu.cn}

\icmlkeywords{Multi-label classification, multi-label performance measures, margin, unified view}

\vskip 0.3in
]



\printAffiliationsAndNotice{}  

\begin{abstract} 
Multi-label classification deals with the problem where each instance is associated with multiple class labels. Because evaluation in multi-label classification is more complicated than single-label setting, a number of performance measures have been proposed. It is noticed that an algorithm usually performs differently on different measures. Therefore, it is important to understand which algorithms perform well on which measure(s) and why. In this paper, we propose a unified margin view to revisit eleven performance measures in multi-label classification. In particular, we define \textit{label-wise} margin and \textit{instance-wise} margin, and prove that through maximizing these margins, different corresponding performance measures are to be optimized. Based on the defined margins, a max-margin approach called LIMO is designed and empirical results validate our theoretical findings.
\end{abstract} 

\section{Introduction} \label{sec:intro}
Multi-label classification aims to build classification models for objects assigned with \textit{multiple} labels simultaneously, which is a common learning paradigm in real-world applications. In text categorization, a document may be associated with a range of topics, such as \textit{science}, \textit{entertainment}, and \textit{news} \cite{schapire2000boostexter}; in image classification, an image can have both \textit{field} and \textit{mountain} tags \cite{boutell2004learning}; in music information retrieval, a piece of music can convey various messages such as \textit{classic}, \textit{piano} and \textit{passionate} \cite{TurnbullBTL08}.

In traditional supervised classification, generalization performance of the learning system is usually evaluated by accuracy, or F-measure if misclassification costs are unequal. In contrast to single-label classification, performance evaluation in multi-label classification is more complicated, as each instance can be associated with multiple labels simultaneously. For example, it is difficult to tell which mistake of the following two cases is more serious: one instance with three incorrect labels vs. three instances each with one incorrect label. Therefore, a number of performance measures focusing on different aspects have been proposed, such as \textit{Hamming loss}, \textit{ranking loss}, \textit{one-error}, \textit{average precision}, \textit{coverage} \cite{schapire2000boostexter}, \textit{micro-F1} and \textit{macro-F1} \cite{tsoumakas2011random}.

Multi-label learning algorithms usually perform differently on different measures; however, there are only a few studies about multi-label performance measures. \citet{dembczynski2010regret} showed that Hamming loss and subset 0/1 loss could not be optimized at the same time. \citet{gao2013consistency} proposed to study the Bayes consistency of surrogate losses for multi-label learning; they proved that none of convex surrogate loss is consistent with ranking loss, and gave a consistent surrogate loss function for Hamming loss in deterministic case. There are a number of studies about F-measure, mostly focusing on single-label tasks, including multi-label learning as application. For example, \citet{NanCLC12} gave justifications and connections about F-measure optimization using decision theoretic approaches (DTA) and empirical utility maximization approaches (EUM). Later, \citet{WaegemanDJCH14} studied the F-measure optimality of inference algorithms from the DTA perspective. \citet{KoyejoNRD15} devoted to study of EUM optimal multi-label classifiers. These theoretical studies offer much insight, though lacking a unified understanding of relation among a variety of multi-label performance measures. Moreover, some performance measures which have been popularly used in evaluation \cite{zhang2015lift} have not been theoretically studied.

In this paper, we try to disclose some shared properties among different measures and establish a unified understanding for multi-label performance evaluation. We propose a \textit{margin} view to revisit eleven commonly used multi-label performance measures, including Hamming loss, ranking loss, one-error, coverage, average precision, macro-, micro- and instance-averaging F-measures and AUCs. Specifically, we propose the concepts of \textit{label-wise} margin and \textit{instance-wise} margin, based on which the corresponding \textit{effectiveness} of multi-label classifiers is defined and then used as bridge to connect different performance measures. Our theoretical results show that by maximizing instance-wise margin, macro-AUC, macro-F1 and Hamming loss are to be optimized, whereas by maximizing label-wise margin, the other eight performance measures except micro-AUC are to be optimized. Inspired by the theoretical findings, we design the LIMO (Label-wise and Instance-wise Margins Optimization) approach to maximize both the two margins. Experiments validate our theoretical findings and demonstrate a flexible way to optimize different measures through one approach by different parameter settings.

The rest of the paper is organized as follows. Section \ref{sec:pre} introduces the notation and definitions of eleven multi-label performance measures. Section \ref{sec:two_margin} proposes the label-wise and instance-wise margins, and presents our theoretical results. Section \ref{sec:approach} presents the LIMO approach. Section \ref{sec:exp} reports the results of experiments. Finally, Section \ref{sec:conclusion} concludes and indicates several future issues.
\vspace{0.3cm}

\section{Preliminaries} \label{sec:pre}

\subsection{Notation}
Assume that $\x_i \in \mathbb{R}^{d\times 1}$ is a real value instance vector, $\y_i \in \{0,1\}^{l\times 1}$ is a label vector for $\x_i$. $m$ denotes the number of training samples. Therefore $y_{ij}\ (i\in \{1,\dots,m\},j\in \{1,\dots,l\})$ means the $j$th label of the $i$th instance, and $y_{ij}=1$ or $0$ means the $j$th label is relevant or irrelevant. The instance matrix is $\X\in \mathbb{R}^{m\times d}$ and the label matrix is $\Y\in \{0,1\}^{m\times l}$. $H: \mathbb{R}^d \rightarrow \{0,1\}^l $ is the multi-label classifier, which consists of $l$ models, one for a label, so $H=\{h_1,\dots,h_l\}$ and $h_j(\x_i)$ denotes the prediction of $y_{ij}$. Moreover, $F: \mathbb{R}^d \rightarrow \mathbb{R}^l$ is the multi-label predictor and the predicted value can be regarded as the confidence of relevance. Similarly, $F$ can be decomposed as $\{f_1,\dots,f_l\}$ where $f_j(\x_i)$ denotes the predicted value of $y_{ij}$.

\vspace{0.1cm}

$H$ can be induced from $F$ via thresholding functions. For example, $h_j(\x_i)=\llbracket f_j(\x_i)>t(\x_i) \rrbracket $ uses a thresholding function based on the instance $\x_i$ and outputs 1 if predicted value is higher than the threshold. $\llbracket \pi \rrbracket$ returns 1 if predicate $\pi$ holds, and 0 otherwise.

\vspace{0.1cm}

For simplification, we use $\Y_{i \cdot}$ to denote the $i$th row vector and $\Y_{\cdot j}$ to denote the $j$th column vector of the label matrix. Furthermore, $Y_{i \cdot}^+$ (or $Y_{i \cdot}^-$) denotes the index set of relevant (or irrelevant) labels of $\Y_{i \cdot}$. Formally, $Y_{i \cdot}^+=\{j \ |y_{ij}=1\}$ and $Y_{i \cdot}^-=\{j \ |y_{ij}=0\}$. In terms of $j$th column of label matrix, $Y_{\cdot j}^+=\{i \ |y_{ij}=1\}$ denotes the index set of positive instances of the $j$th label and $Y_{\cdot j}^-=\{i \ |y_{ij}=0\}$ denotes the set of negative instances similarly. We use $|\cdot|$ to denote the cardinality of a set, thus, the number of relevant labels of $\x_i$ is $|Y_{i \cdot}^+|$.

\subsection{Multi-label Performance Measures} \label{ss:measures}
Table \ref{table:measures} summarizes the eleven multi-label performance measures commonly used in previous studies. The first five measures (Hamming loss, ranking loss, one-error, coverage, average precision) are considered in \citet{schapire2000boostexter} and a multitude of works, e.g., \citet{HuangYZ12} and \citet{zhang2015lift}. The next six measures are extensions of F-measure and AUC (the Area Under the ROC Curve) in multi-label classification via different averaging strategies. These F-measures are popluar both in algorithm evaluation \cite{LiuT15} and theoretical analysis \cite{KoyejoNRD15}. AUCs are used for algorithm evaluation such as in \citet{Lampert11}, \citet{PhamRFA15} and \citet{zhang2015lift}. 

\vspace{0.1cm}

Some of these measures are defined on classifier $H$, and they care about the binary classification performance. While some of these measures are defined on predictor $F$, and they usually measure the ranking performance of the predictor. We have noticed that some performance measures on ranking are ill-defined when $F$ is a constant function. For example, if $F$ outputs 1 for all labels, then \textit{one-error}$(F)=0$, $\textit{coverage}(F)=0$ and various AUCs will be 1, which are the optimal values respectively. In multi-label learning community, there is often an underlying assumption that a total ranking can be induced from continuous real-value predictions, which is common in practical cases. In this paper, we still stick to the convention in previous works and assume that no tie happens in continuous prediction to solve this definition flaw. 

\begin{table*}[!htb]
\centering
\caption{Definitions of eleven multi-label performance measures} \label{table:measures}
\scriptsize
\renewcommand{\arraystretch}{0.8}
\setlength{\abovedisplayskip}{0pt}
\setlength{\belowdisplayskip}{0pt}
\setlength{\abovedisplayshortskip}{0pt}
\setlength{\belowdisplayshortskip}{0pt}  
\begin{tabular}{M{2.6cm} | M{8cm} | L{4.2cm} } 
\hline
\vspace{0.1cm}
 \small \textbf{Measure} & \small \textbf{Formulation} &  \small \makecell[c]{\textbf{Note}} \\
\hline
 \small Hamming loss & \begin{equation*}
  hloss(H)=\frac{1}{m l} \sum_{i=1}^{m} \sum_{j=1}^{l} \llbracket h_{ij}\neq y_{ij} \rrbracket
\end{equation*} & \fontsize{8}{8}\selectfont The fraction of misclassified labels\\
\hline
\small  ranking loss & \begin{equation*}
  rloss(F)=\frac{1}{m}\sum_{i=1}^m \frac{|\mathcal{S}_{\text{rank}}^i|}{|Y_{i\cdot}^+||Y_{i\cdot}^-|} 
\end{equation*}
\begin{equation*}
  \mathcal{S}_{\text{rank}}^i = \{ (u,v)| f_u(\x_i)\leq f_v(\x_i), (u,v)\in Y_{i\cdot}^+\times Y_{i\cdot}^- \}
\end{equation*} &  \fontsize{8}{8}\selectfont  The average fraction of reversely ordered label pairs of each instance.\\
\hline
\small one-error & \begin{equation*}
  one\text{-}error(F)=\frac{1}{m}\sum_{i=1}^m \llbracket \argmax F(\x_i) \notin Y_{i\cdot}^+ \rrbracket
\end{equation*} & \fontsize{8}{8}\selectfont The fraction of instances whose most confident label is irrelevant.\\
\hline
\small coverage & \begin{equation*}
  coverage(F)=\frac{1}{m}\sum_{i=1}^m \llbracket \max_{\ j\in Y_{i\cdot}^+} rank_F(\x_i,j)-1 \rrbracket
\end{equation*} & \fontsize{8}{8}\selectfont The number of more labels on average should include to cover all relevant labels\\
\hline
\small average precision & \begin{equation*}
    avgprec(F)=\frac{1}{m}\sum_{i=1}^m \frac{1}{|Y_{i\cdot}^+|} \sum_{j\in Y_{i\cdot}^+} \frac{|\mathcal{S}_{\text{precision}}^{ij}|}{rank_F(\x_i,j)} 
    \end{equation*}
    \begin{equation*}
    \mathcal{S}_{\text{precision}}^{ij}=\{k\in Y_{i\cdot}^+ | rank_F(\x_i,k)\leq rank_F(\x_i,j) \}
\end{equation*} &\fontsize{8}{8}\selectfont  The average fraction of relevant labels ranked higher than one other relevant label.\\
\hline  
\small macro-F1 & \begin{equation*}
  \text{\textit{macro-F1}}(H)=\frac{1}{l}\sum_{j=1}^l \frac{2 \sum_{i=1}^m y_{ij} h_{ij}}{\sum_{i=1}^m y_{ij}+ \sum_{i=1}^m h_{ij}}
\end{equation*} & \fontsize{8}{8}\selectfont F-measure averaging on each label.\\
\hline
\small instance-F1 & \begin{equation*}
  \text{\textit{instance-F1}}(H)= \frac{1}{m} \sum_{i=1}^{m} \frac{2 \sum_{j=1}^l y_{ij} h_{ij}} { \sum_{j=1}^l y_{ij}+ \sum_{j=1}^l h_{ij} }
\end{equation*} & \fontsize{7.8}{8}\selectfont F-measure averaging on each instance.\\
\hline
\small micro-F1 & \begin{equation*}
  \text{\textit{micro-F1}}(H)= \frac{2 \sum_{j=1}^l \sum_{i=1}^m y_{ij} h_{ij}} {\sum_{j=1}^l \sum_{i=1}^m y_{ij}+ \sum_{j=1}^l \sum_{i=1}^m h_{ij}}
\end{equation*} & \fontsize{8}{8}\selectfont F-measure averaging on the prediction matrix.\\
\hline
\small macro-AUC & \begin{equation*}
  \text{\textit{macro-AUC}}(F) = \frac{1}{l} \sum_{j=1}^{l} \frac{|\mathcal{S}_{\text{macro}}^j|}{|Y_{\cdot j}^+||Y_{\cdot j}^-|}
\end{equation*}
\begin{equation*}
  \mathcal{S}_{\text{macro}}^j = \{ (a,b) \in Y_{\cdot j}^+ \times Y_{\cdot j}^- | f_j(\x_a)\geq f_j(\x_b)\}
\end{equation*} & \fontsize{8}{8}\selectfont AUC averaging on each label. $\mathcal{S}_{\text{macro}}$ is the set of correctly ordered instance pairs on each label.\\
\hline
\small instance-AUC & \begin{equation*}
  \text{\textit{instance-AUC}}(F) = \frac{1}{m} \sum_{i=1}^{m} \frac{|\mathcal{S}_{\text{instance}}^i|}{|Y_{i\cdot}^+||Y_{i\cdot }^-|}
\end{equation*}
\begin{equation*}
  \mathcal{S}_{\text{instance}}^i = \{ (u,v) \in Y_{i \cdot }^+ \times Y_{i \cdot }^- | f_u(\x_i)\geq f_v(\x_i)\}
\end{equation*} & \fontsize{8}{8}\selectfont AUC averaging on each instance. $\mathcal{S}_{\text{instance}}$ is the set of correctly ordered label pairs on each instance.\\
\hline
\small micro-AUC & \begin{equation*}
  \text{\textit{micro-AUC}}(F) =\frac{|\mathcal{S}_{\text{micro}}|}{(\sum_{i=1}^m |Y_{i\cdot}^+|)\cdot(\sum_{i=1}^m |Y_{i\cdot }^-|)}
\end{equation*}
\begin{equation*}
  \mathcal{S}_{\text{micro}} = \{ (a,b,i,j)  |(a,b) \in Y_{ \cdot i }^+ \times Y_{\cdot j}^-,\  f_i(\x_a)\geq f_j(\x_b)\}
\end{equation*} & \fontsize{8}{8}\selectfont AUC averaging on prediction matrix. $\mathcal{S}_{\text{micro}}$ is the set of correct quadruples.\\
\hline
\end{tabular}
\end{table*}
\section{Theoretical Results} \label{sec:two_margin}
Here we define two new concepts: label-wise margin and instance-wise margin.

\begin{mydef} Given a multi-label predictor $F: \mathbb{R}^d \rightarrow \mathbb{R}^l$ and $F=\{f_1,\dots,f_l\}$, a training set $(\X,\Y)$, the \textbf{label-wise margin} on instance $\x_i$ is defined as:
\begin{equation*}
  \gamma^{\text{label}}_i= \min_{u,v} \{f_u(\x_i)-f_v(\x_i) \ |\ (u,v)\in Y_{i \cdot}^+ \times Y_{i \cdot}^- \}.
\end{equation*}
\end{mydef}
\vspace{-0.2cm}
$Y_{i \cdot}^+ \times Y_{i \cdot}^-$ is the set of all the (relevant, irrelevant) label index pairs of instance $i$.
\begin{mydef} Given a multi-label predictor $F: \mathbb{R}^d \rightarrow \mathbb{R}^l$ and $F=\{f_1,\dots,f_l\}$, a training set $(\X,\Y)$, the \textbf{instance-wise margin} on label $\Y_{\cdot j}$ is defined as:
\begin{equation*}
   \gamma^{\text{inst}}_j=\min_{a,b} \{f_j(\x_a)-f_j(\x_b) \ |\  (a,b)\in Y_{\cdot j }^+ \times Y_{\cdot j }^-  \}.
\end{equation*}
\end{mydef}
\vspace{-0.2cm}
 $Y_{\cdot j}^+ \times Y_{\cdot j}^-$ is the set of all the (positive, negative) instance index pairs of label $j$.

Label-wise margin and instance-wise margin describe the discriminative ability of $F$. The larger the label-wise margin, the easier to distinguish relevant and irrelevant labels of an instance. Meanwhile, the larger the instance-wise margin, the easier for $F$ to distinguish positive and negative instances of a particular label. Therefore, we want to maximize label-wise/instance-wise margin to get better performance.

Although we prefer maximizing these two margins, with respect to performance measures, the objective can be relaxed. We define three properties a predictor $F$ can have: label-wise effective, instance-wise effective and double effective.

\begin{mydef}
If all the label-wise margins of $F$ on a dataset $D=(\X,\Y)$ are positive, this predictor $F$ is \textbf{label-wise effective} on $D$.
\end{mydef}

\begin{mydef}
If all the instance-wise margins of $F$ on a dataset $D=(\X,\Y)$ are positive, this predictor $F$ is \textbf{instance-wise effective} on $D$.
\end{mydef}

\begin{mydef}
If all the label-wise margins \textbf{and} instance-wise margins of $F$ on a dataset $D=(\X,\Y)$ are positive, this predictor $F$ is \textbf{double effective} on $D$.
\end{mydef}
\vspace{-0.1cm}
Roughly speaking, label-wise effective means $F$ can exactly distinguish relevant and irrelevant labels of each instance and instance-wise effective means $F$ can exactly distinguish positive and negative instances of every label. Not surprisingly, double effective $F$ has the strongest ability in distinguishing.

In the next two subsections, we use the effectiveness to analyze different performance measures, and summarize the analysis results in Section \ref{ss:summary}.

\subsection{Performance Measures on Ranking} \label{ss:measures_f}
Several multi-label performance measures can be empirically optimized according to the following theorems:

\begin{mythm} 
If a multi-label predictor $F$ is label-wise effective on $D$, then ranking loss, one-error, coverage, average precision and instance-AUC are optimized on the dataset.
\end{mythm}

\begin{spacelessprf}
\setdefaultleftmargin{0pt}{}{}{}{}{}
\setdefaultenum{(a)}{}{}{}
\begin{enumerate}
\item Ranking loss: From the definition of label-wise effective, for every pair $(u,v)\in Y_{i \cdot}^+ \times Y_{i \cdot}^-$, we have $f_u(\x_i)>f_v(\x_i)$. Therefore, the reversed set $\mathcal{S}_{\text{rank}}^i$ (in Table \ref{table:measures} ranking loss) is empty and the cardinality of the set is zero, which implies the cardinality sum of all reversed sets $rloss(F)=0$. Ranking loss is optimized.
\item One-error: For a label-wise effective $F$, because label-wise margin is positive on an instance $\x_i$, we have:
\begin{equation*}
\max_u f_u(\x_i)> \max_v f_v(\x_i), \forall u\in Y_{i \cdot}^+ , \forall v \in  Y_{i \cdot}^- .
\end{equation*}
Then
\begin{equation*}
  \forall \x_i, \ \argmax F(\x_i) \in Y_{i \cdot}^+ .
\end{equation*}
Thus, $\llbracket \argmax F(\x_i) \notin Y_{i\cdot}^+ \rrbracket=0$ for every instance $\x_i$, and $one \text{-}error(F)=0$. One-error is optimized.
\item Coverage: When $F$ is label-wise effective, the maximum rank of a relevant label is less than the minimum rank of an irrelevant label, which means:
\begin{align}
  \max_{u\in Y_{i \cdot}^+} rank_F (\x_i,u) &< \min_{v\in Y_{i \cdot}^-} rank_F(\x_i,v), \label{eq:maxmin}\\
  \max_{u\in Y_{i \cdot}^+} rank_F (\x_i,u) &= |Y_{i \cdot}^+|. \nonumber
\end{align}  
Therefore, coverage can be calculated as:
\begin{equation*}
  coverage(F) = \frac{1}{m} \sum\nolimits ^{m}_{i=1} [|Y_{i \cdot}^+|-1 ].
\end{equation*}
Which is the optimal value of coverage.
\item Average precision: Assume that $j$ is a relevant label of instance $i$, it follows from Equation (\ref{eq:maxmin}) that:
\begin{equation*}
rank_F(\x_i,j) = |\{k\in Y_{i\cdot}^+ | rank_F(\x_i,k)\leq rank_F(\x_i,j) \}| 
\end{equation*}
Since $rank_F(\x_i,j)$ is exactly the definition of $\mathcal{S}_{\text{precision}}^{ij}$, $avgprec(F)=1$, i.e, average precision is optimized.\\
\item Instance-AUC: Because of label-wise effective, for an instance $\x_i$, we have:
\begin{equation*}
   f_u(\x_i)> f_v(\x_i), \forall (u,v) \in Y_{i \cdot }^+ \times Y_{i \cdot }^- .
\end{equation*}
Therefore, the size of the correct ordered prediction value pair on instance $i$ is:
\begin{equation*}
  |\{ (u,v) \in Y_{i \cdot }^+ \times Y_{i \cdot }^- | f_u(\x_i)\geq f_v(\x_i)\}| = |Y_{i \cdot }^+||Y_{i \cdot}^-|. 
\end{equation*}
So \textit{instance-AUC}$(F)=1$ and instance-AUC is optimized.
\end{enumerate}

\end{spacelessprf}
Similar to the proof of instance-AUC, we can prove the result of macro-AUC:
\begin{mythm}
If a multi-label predictor $F$ is instance-wise effective on $D$, then macro-AUC is optimized.
\end{mythm}
\begin{spacelessprf}
Because of instance-wise effective, for a label vector $\Y_{\cdot j}$, we have:
\begin{equation*}
   f_j(\x_a)> f_j(\x_b), \forall (a,b) \in Y_{\cdot j}^+ \times Y_{ \cdot j}^- .
\end{equation*}
Therefore, the size of the correct ordered prediction value pair on label $j$ is:
\begin{equation*}
  \{ (a,b) \in Y_{\cdot j}^+ \times Y_{\cdot j}^- | f_j(\x_a)\geq f_j(\x_b)\} = |Y_{\cdot j }^+||Y_{\cdot j}^-| .
\end{equation*}
So \textit{macro-AUC}$(F)=1$ and macro-AUC is optimized.
\end{spacelessprf}
Micro-AUC sees the label matrix as a whole and cannot be optimized by instance-wise effective $F$ or label-wise effective $F$. However, the double effective $F$ is much more powerful. We now prove the following result of micro-AUC. 
\begin{mythm} \label{th:micro-AUC}
If a multi-label predictor $F$ is double effective on $D$, then as the number of instances grows, \textit{micro-AUC} is optimized.
\end{mythm}
\begin{spacelessprf}
 We first prove a result of random variables $A_i,B,C$. If $n$ random variables $A_1,A_2,\cdots,A_n$ are drawn from uniform distribution $U(0,1)$, for a random constant $a$, the event that at least one $A_i$ is smaller than $a$ is:
\begin{equation*}
  \Pr[\exists A_i,\ A_i \leq a] = 1-(1-a)^n .
\end{equation*}
Another random variable $B$ is uniformly distributed in $(0,\min\{A_i\})$, and the probability that a random variable $C\sim U(0,1)$ is bigger than $B$ is:
\begin{align}
 \Pr[C>B ] \geq& \Pr[(C \geq a) \wedge (\exists A_i,\ A_i \leq a)]\nonumber \\
            =& (1-\frac{a}{2})[1-(1-a)^n] . \label{lower_bound}
\end{align}
For any small $a$, we can choose a large enough $n$ to make Equation (\ref{lower_bound}) close to 1.

Given a label matrix $\Y \in \{0,1 \}^{m\times l}$ and the corresponding prediction matrix $\F \in (0,1)^{m\times l}$, because predictor $F$ is double effective, the prediction matrix satisfies the following conditions:
\begin{align*}
    F_{ij}&>F_{iu} \text{ if } Y_{ij}=1\wedge Y_{iu}=0,\\
    F_{ij}&>F_{vj} \text{ if } Y_{ij}=1\wedge Y_{vj}=0.
\end{align*}
To force the value in $\F$ is in $(0,1)$, we further assume a uniform distribution $F_{ij}\sim U(0,1)$ when $Y_{ij}=1$. 

If $Y_{ij}=0$, then $F_{ij}$ should be less than $F_{iu}$ if $Y_{iu}=1$ and $F_{vj}$ if $Y_{vj}=1$. Suppose that the minimum value $b$ is defined as:
\begin{equation*}
      b=\min\Big\{ \min\limits_{v} \{F_{vj}| Y_{vj}=1\},\min\limits_{u} \{F_{iu}| Y_{iu}=1\}\Big\}.
\end{equation*}
Then $F_{ij}$ is drawn from $U(0,b)$. And we can choose a small constant value $a>b$. 

According to Equation (\ref{lower_bound}), the probability that a random pair $(i,j,u,v)$ to be a correct micro pair is:
\begin{align*}
  P_\text{micro}&=\Pr[ F_{ij}>F_{uv} | Y_{ij}=1,Y_{uv}=0] \\
  &\geq (1-\frac{a}{2})[1-(1-a)^n],\\
  \text{where } n&=\frac{k}{ml}(m+l-2)
\end{align*}
In the practical case, the number of labels is proportional to the number of instances: $k \propto m$. We assume $k=pm$ where $p$ is a constant smaller than $l$.
\begin{align*}
\lim_{m\to \infty} &n = \lim_{m\to \infty} \frac{p}{l}(m+l-2) = \infty,\\
\lim_{m\to \infty} & \frac{|\mathcal{S}_{\text{micro}}|}{|(\sum_{i=1}^m |Y_{i\cdot}^+|)\cdot(\sum_{i=1}^m |Y_{i\cdot }^-|)|}= \lim_{m\to \infty}  P_\text{micro}=1 .
\end{align*}
Therefore, micro-AUC is to be optimized as the number of instances grows.
\end{spacelessprf}

With the above analysis, we can conclude that a label-wise effective $F$ can optimize ranking loss, one-error, coverage, average precision, instance-AUC, micro-AUC and an instance-wise effective $F$ can optimize macro-AUC. For micro-AUC, a double effective $F$ can optimize it as the number of instances increases.

\subsection{Performance Measures on Classification} \label{ss:measures_h}
As mentioned in Section \ref{ss:measures}, there are some measures evaluating classifier $H$ instead of predictor $F$. There are many thresholding or binarization strategies \cite{fan2007study,furnkranz2008multilabel,ClassifierChains}. For simplicity, we focus on two main strategies: thresholding on each instance and thresholding on each label.

A label-wise effective $F$ can be equipped with a thresholding function based on each instance such as $t(\x_i)$ and construct the $H$ by $h_j(\x_i)=\llbracket f_j(\x_i)>t(\x_i) \rrbracket$. However, using $t(\x_i)$ on an instance-wise effective $F$ is unreasonable since the predicted values on different labels may not be comparable. In a word, we should use suitable threshold function on different effective $F$s, i.e., $t(\x_i)$ on each instance for label-wise effective $F$, and $t_j$ on each label for instance-wise effective $F$. It is reasonable to use either $t(\x_i)$ or $t_j$ for double effective $F$. 

To formally analyze the performance measures on classification, we define the threshold error:
\begin{mydef}
Given a descending ordered real-value sequence $x_1,x_2,\dots,x_k$ with an optimal cut number $c^*$, where $c^*\in \mathbb{N}$ and $1 \leq c^*\leq k$. For a real value threshold $t\in (x_k-1, x_1+1)$, the \textbf{threshold error} $\epsilon=|\argmin_i (x_i)-c^*|$ where $x_i>t$.
\end{mydef}
Intuitively, the threshold error $\epsilon$ counts how many items are incorrectly classified on a descending ordered sequence where the correct answer is $c^*$. Based on the threshold error, we propose the following theorems about performance measures on classification.
\begin{mythm} \label{th:label_th}
For a label-wise effective $F$, if the thresholding function makes at most $\epsilon_i$ error on each instance $i$, the micro-F1, instance-F1 and Hamming loss are bounded as follows:
\begin{align*}
   \text{\textit{micro-F1}}(H)&= \text{\textit{instance-F1}}(H)\\
    &\geq \frac{1}{m} \sum_{i=1}^m \min \Big\{\frac{2(|Y_{i \cdot}^+|-\epsilon_i)}{2|Y_{i \cdot}^+|-\epsilon_i}, \frac{2|Y_{i \cdot}^+|}{2|Y_{i \cdot}^+|+\epsilon_i}\Big\},\\  
    \text{\textit{hloss}}(H) &\leq \frac{1}{m l}\sum \nolimits_{i=1}^{m} \epsilon_i   .
\end{align*}
\end{mythm}

The main idea of the above theorem is that, given the threshold error and the number of relevant labels, we can compute the gap between the worst possible and the perfect contingency table. Hence the F-measure is based on the contingency table, the lower bound can be deduced. The detailed proof of Theorem \ref{th:label_th} is in Appendix A.1.

Similar to Theorem \ref{th:label_th}, we can prove the results for label-wise effective $F$:
\begin{mythm} \label{th:inst_th}
For an instance-wise effective $F$, if the thresholding function makes at most $\epsilon_j$ error on each label $j$, then the macro-F1 and Hamming loss are bounded as follows:
\begin{align*}
    \text{\textit{macro-F1}}(H)&\geq \frac{1}{l} \sum_{j=1}^l \min \Big\{\frac{2(|Y_{\cdot j}^+|-\epsilon_j)}{2|Y_{\cdot j}^+|-\epsilon_j}, \frac{2|Y_{\cdot j}^+|}{2|Y_{\cdot j}^+|+\epsilon_j}\Big\}
    ,\\
    \text{\textit{hloss}}(H)&\leq \frac{1}{ml}\sum \nolimits_{j=1}^{l} \epsilon_j .
\end{align*}
\end{mythm}
The detailed proof of Theorem \ref{th:inst_th} is in Appendix A.2.

With the above analysis, we can conclude that a label-wise effective $F$ can optimize instance-F1 and micro-F1, an instance-wise effective $F$ can optimize macro-F1. Both the two effective $F$s can optimize Hamming loss. For a double effective $F$, because it enjoys both the properties, it can optimize all the above mentioned performance measures if proper thresholds are used.

\subsection{Summary} \label{ss:summary}

Table \ref{table:category} summarizes our theoretical results in Section \ref{ss:measures_f} and \ref{ss:measures_h}. Each row shows the results of one multi-label performance measure. Note that double effective is a special case of label-wise effective and instance-wise effective and thus, if one performance measure is optimized by either label-wise or instance-wise effective predictor, it will also be optimized by double effective predictor.

\begin{table}[!tb]
\centering
\setlength{\tabcolsep}{3pt}
\renewcommand{\arraystretch}{1.1}
\caption{Summary of performance measures optimized by $x$-effective multi-label predictor ($F$). `\checkmark' means $F$ in this cell is proved to optimize this measure; `\xmark' means $F$ in this cell does not necessarily optimize the measure; `$\bullet$'/`$\circ$' means the calculation is with/without thresholding.} \label{table:category}
\vspace{0.1cm}
\begin{tabular}{ c | c | c |c | c } 
\hline
\multirow{2}{*}{Measure} & \multicolumn{3}{c|}{$x$-effective $F$} & \multirow{2}{*}{Threshold} \\
\cline{2-4}
  & label-wise & inst-wise & double & \\
\hline
ranking loss & \checkmark & \xmark & \checkmark & $\circ$ \\
avg. precision & \checkmark & \xmark & \checkmark & $\circ$ \\
one-error   &\checkmark & \xmark & \checkmark & $\circ$ \\
coverage & \checkmark & \xmark & \checkmark & $\circ$ \\
instance-AUC &\checkmark & \xmark & \checkmark & $\circ$ \\
macro-AUC & \xmark & \checkmark & \checkmark & $\circ$ \\
micro-AUC & \xmark & \xmark & \checkmark & $\circ$ \\
\hline
macro-F1 & \xmark & \checkmark & \checkmark & $\bullet$ \\
instance-F1 & \checkmark & \xmark & \checkmark & $\bullet$ \\
micro-F1 & \checkmark & \xmark & \checkmark & $\bullet$ \\
Hamming loss & \checkmark & \checkmark & \checkmark & $\bullet$ \\
\hline
\end{tabular}
\end{table}

In the light of the analysis, the performance on different performance measures through optimizing margins can be expected. For example, if one maximizes instance-wise margin on each label, s/he will get good performance on macro-AUC but may suffer higher loss on ranking loss, coverage and some other measures where `\xmark' marked in the inst-wise column. If one tries to maximize the label-wise margin but pay no attention to instance-wise margin, s/he may perform well on average precision but poor on macro-F1 (e.g., \citet{elisseeff2002kernel}). Maximzing both the label-wise margin and instance-wise margins to get a double effective $F$ is expected to be the best choice.


\section{The LIMO Approach} \label{sec:approach}
The above analysis reveals that maximizing different margins will optimize different measures, and if possible, double effective $F$ is prefered since it enjoys the benefits of maximizing both the label-wise margin and the instance-wise margin. Therefore, we propose the LIMO approach. LIMO is a single approach which can optimize both the two margins, and it can also be degenerated to optimize either margin seperately via parameter setting.

\subsection{Formulation} \label{ss:proposed_formulation}
Suppose that $F$ is a linear predictor, which means $F(\X)=\W^T \X$ where $\W=[\w_1,\w_2,\cdots,\w_l]$. We propose the following formulation:

 \setlength{\belowdisplayskip}{0pt} \setlength{\belowdisplayshortskip}{0pt}
 \setlength{\abovedisplayskip}{0pt} \setlength{\abovedisplayshortskip}{0pt}
\begin{equation} \label{obj}
 \begin{split}
     \argmin_{\W, \xi} &\ \sum_{i=1}^l ||\w_i||^2+\lambda_1 \sum_{i=1}^m \sum_{(u,v)}\xi_i^{uv}+\lambda_2 \sum_{j=1}^l \sum_{(a,b)} \xi_{ab}^j \\
     \text{s.t.}\ & \w_u^\top \x_i -\w_v^\top \x_i >1-\xi_i^{uv}, \ \ \xi_i^{uv}\geq 0, \\
     & \quad\text{for }i=1,\cdots,m \text{ and } (u,v)\in Y_{i \cdot}^+ \times Y_{i \cdot}^- \ , \\
                &\w_j^\top \x_a -\w_j^\top \x_b>1-\xi^j_{ab}, \  \ \xi^j_{ab}\geq 0, \\
    & \quad\text{for } j=1,\cdots,l \text{ and } (a,b)\in Y_{\cdot j}^+ \times Y_{\cdot j}^- \ .
\end{split}    
\end{equation} 
Here $\xi_i^{uv}$ and $\xi^j_{ab}$ are the slack variables, and $\lambda_1,\lambda_2$ are the trade-off parameters. When both $\lambda_1$ and $\lambda_2$ are positive, both label-wise and instance-wise margins are considered. If we set $\lambda_1=0$ (or $\lambda_2=0$), then only the instance-wise (or label-wise) margin is considered. In this paper, if the approach only considers instance-wise (or label-wise) margin , we call the approach as LIMO-inst (or LIMO-label). And LIMO considers both the two margins.

\begin{algorithm}[!htb]
  \caption{LIMO}\label{alg:limo}
\begin{algorithmic}[1]
\INPUT~~\\
    Data matrix $\X \in \mathbb{R}^{m\times d}$, label matrix $\Y\in \{0,1\}^{m\times l}$, \\
    step size $\eta$, trade-off parameters $\lambda_1, \lambda_2$, and the maximium iteration number $T$.
\PROCEDURE~~\\
    \STATE Initialize $\W^0$ with $N(0,1/\sqrt{d})$ random values.
    \STATE Compute the weight vector $\bc^{inst}$ of each instance, $\bc^{inst}_i=|Y_{i \cdot}^+ ||Y_{i \cdot}^-|/ \sum_{i=1}^m |Y_{i \cdot}^+ ||Y_{i \cdot}^-|$.
      \STATE Compute the weight vector $\bc^{label}$ of each label, $\bc^{label}_j=|Y_{ \cdot j}^+ ||Y_{ \cdot j}^-| / \sum_{j=1}^l |Y_{ \cdot j}^+ ||Y_{\cdot j}^-|$.
    \FOR{$t=1,2,\cdots,T$}
      \STATE Random sample an instance $\x^t_i$ using weight $\bc^{inst}$,
      \STATE Random sample a positive label $y_{iu}$ and a negative label $y_{iv}$ of instance $\x^t_i$.
      \IF{ $1-\w_u^\top \x^t_i+\w_v^\top\x^t_i>0$}
          \STATE $\w_u^t=\w_u^{t-1}-\eta(-\lambda_1\x^t_i + \w_u^{t-1})$.
          \STATE $\w_v^t=\w_v^{t-1}-\eta(\lambda_1\x^t_i+ \w_v^{t-1})$.
      \ENDIF

      \STATE Random sample index $j$ of label using weight $\bc^{label}$.
      \STATE Random sample a positive instance $\x^t_a$ and a negative instance $\x^t_b$ on label $j$.
      \IF{ $1-\w_j^\top \x^t_a+\w_j^\top \x^t_b>0$}
          \STATE $\w_j^t=\w_j^{t-1}- \eta( \lambda_2(\x^t_b-\x^t_a) + \w_j^{t-1})$.
      \ENDIF
    \ENDFOR
    \STATE $\W=\frac{1}{T} \sum_{t=1}^{T} \W^t$.

\OUTPUT~~\\
    Multi-label linear model $\W$.
\end{algorithmic}
\end{algorithm} 
\subsection{Algorithm}
The objective Equation (\ref{obj}) is difficult to solve directly because of the large number of constraints and slack variables. For a training set with $m$ instances and $l$ labels, the number of constraints will be $O(m^2 l +ml^2)$, which may exceed memory limit in real-world applicaitons.

In order to deal with the computational problem, we solve Equation (\ref{obj}) by stochastic gradient descent (SGD) with fixed step size and the default averaging technique in \citet[Chapter 14.3]{Shalev-Shwartz}. The key point of SGD is to find out a random vector, whose expected value at each iteration equals the gradient direction. We randomly sample two kinds of triplets and use them to compute the correct direction. At each iteration $t$, we sample a triplet $(\x^t_i, y_{iu}, y_{iv})$ where $y_{iu}$ is relevant and $y_{iv}$ is irrelevant, and a triplet $(j, \x_a^t, \x_b^t)$ where $\x_a^t$ is a positive instance and $\x_b^t$ is a negative instance both on label $j$. Then we use the two triplets to compute the random gradient vector for SGD. The detailed algorithm is presented in Algorithm \ref{alg:limo} and the proof that the random vector is an unbiased estimation of the gradient direction is available in Appendix A.3. 

After the training procedure, we can use the linear model to predict continuous confidence values on the training data, then choose the best threshold value by optimizing a specific classification measure. 

\section{Experiments} \label{sec:exp}
We conduct experiments with LIMO on both synthetic and benchmark data. Note that the main purpose of our work is to study multi-label performance measures from the aspect of margin optimization, and thus, the goal of our experiments is to validate our theoretical findings rather than claim that LIMO is superior, although its performance is really highly competitive.

\subsection{Synthetic Data}
\begin{figure}[h]
\centering
\begin{minipage}[c]{4.6cm}
\vspace{-0.2cm}
    \caption{Input space consists of four regions with different assignments of the label set $\{A, B, C, D\}$. The center point is with coordinate $(0, 0)$.
    } \label{fig:label}
  \end{minipage} 
\begin{minipage}[c]{3cm}
    \includegraphics[width=2.6cm]{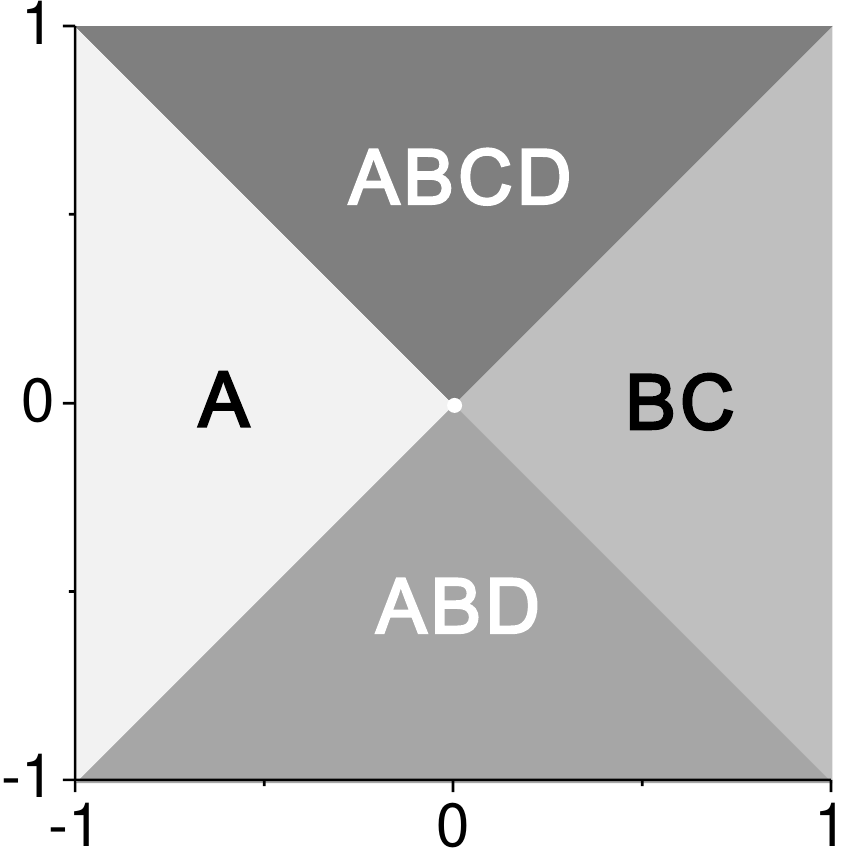}
  \end{minipage}
\end{figure}

We conduct experiments on synthetic data with 4 labels. 2000 data points are randomly generated from a $(-1,+1)^2$ square, and the labels are assigned as in Figure \ref{fig:label}. 50\% data are held out for testing. The synthetic data is designed to simulate a typical real-world circumstance. The number of co-occurrent labels varies, the regions of each label are different and the data cannot be perfectly seperated by a linear learner. 

\begin{figure}[htb] 
 \centering
 \includegraphics[width=8cm]{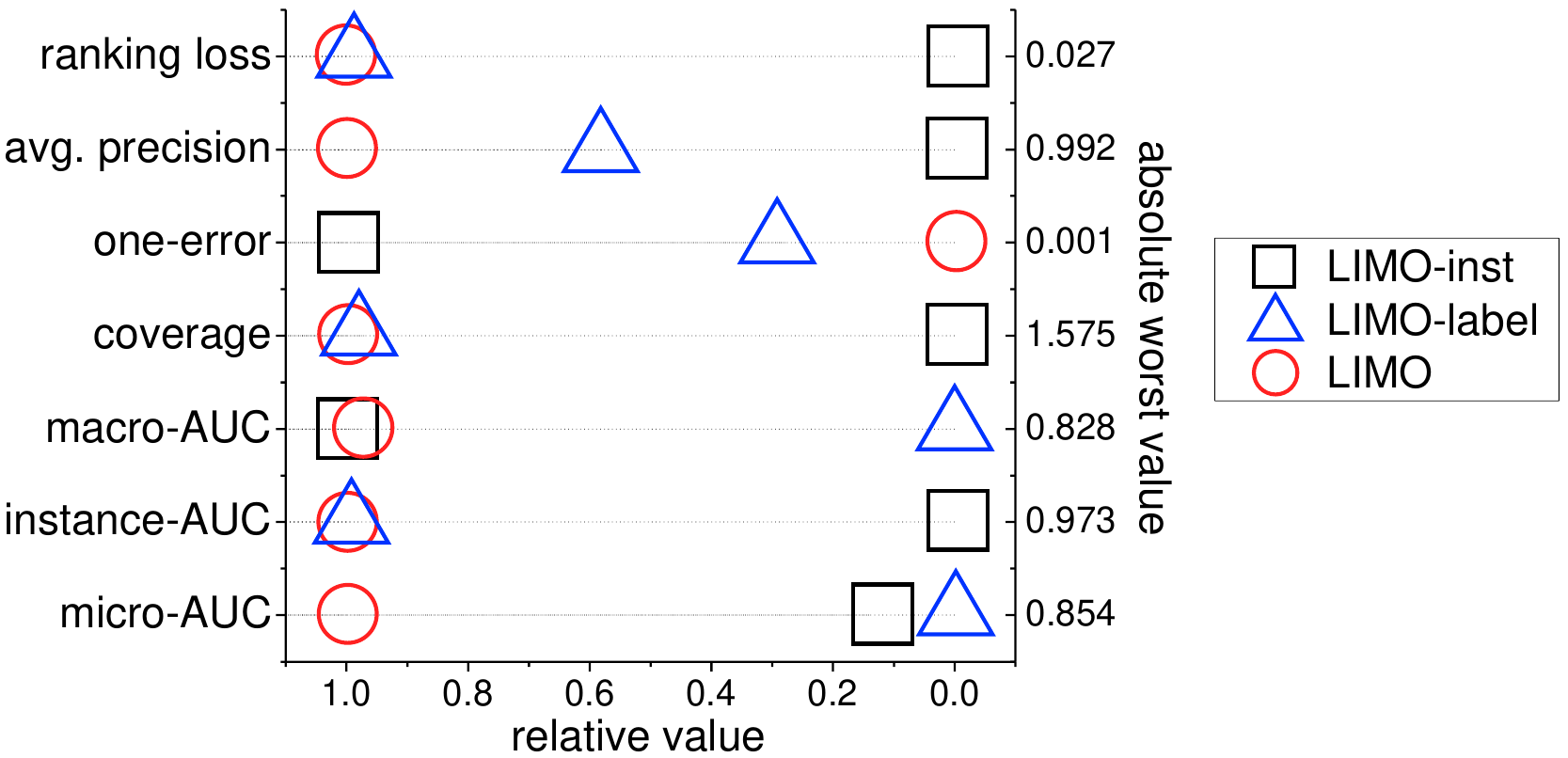} 
 \vspace{-0.4cm}
 \caption{Summary of the relative performance on ranking measures. The more to the left, the better the performance.} \label{fig:f_measures}
\end{figure}

To demonstrate the relationship between margins and performance measures, we degenerate LIMO to only consider either margin by setting the trade-off parameter $\lambda_1$ or $\lambda_2$ to zero. LIMO-inst sets $\lambda_1=0$ and LIMO-label sets $\lambda_2=0$. The other parameter is set to 100 and LIMO sets $\lambda_1=\lambda_2=100$. Ten replications of the experiment are conducted and the average results are reported. Because the range of performance measure coverage is not $[0,1]$, while some performance measures are better when higher, and some are better when lower, we rescale all the performance values into relative values for clearer visualization. The best one is rescaled to 1 and the worst one is rescaled to 0. Figure \ref{fig:f_measures} shows the relative results, where the originally worst performance value is given on the right.

The results shown in Figure \ref{fig:f_measures} support our theoretical findings in Table \ref{table:category}. For example, micro-AUC is considered to be optimized by double effective $F$ but not the other two, therefore LIMO (the red circle) gets the best relative value. For some measures proved to be optimized by label-wise margin such as ranking loss, average precision, coverage and instance-AUC, LIMO-label beats LIMO-inst. While for macro-AUC, LIMO-inst wins. For one-error, all three versions of LIMO do extremely well and get less than 0.001 absolute value, which may be the reason why the relative values are unexpected.

\begin{figure}[htb] 
 \centering
 \includegraphics[width=8cm]{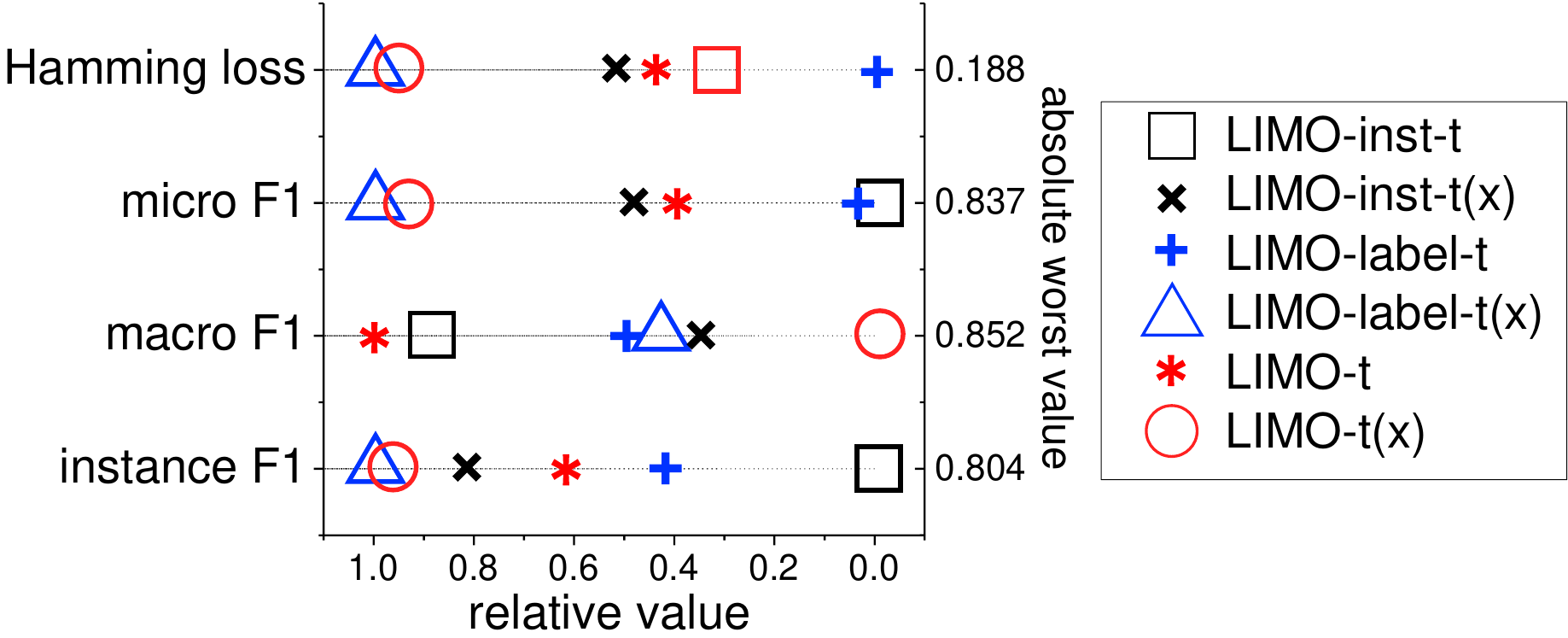} 
 \vspace{-0.4cm}
 \caption{Summary of the relative performance on classification measures. The more to the left, the better the performance.} \label{fig:h_measures}
\end{figure}

Figure \ref{fig:h_measures} shows the relative performance on classification. We use two types of thresholding discussed in Section \ref{ss:measures_h}: threshold function based on each instance or each label (denoted by -t(x) or -t in the legend). The thresholds are estimated on training data. This figure exactly shows our theoretical results: LIMO-label equipped with $t(x)$ can optimize instance-F1 and micro-F1; LIMO-inst equipped with $t$ can optimize macro-F1. By considering both label-wise margin and instance-wise margin, LIMO works well on all four classificaiton measures.

\subsection{Benchmark Data}
We conduct experiments on eleven multi-label performance measures to further show that optimizing the label-wise or the instance-wise margin can lead to different results, as revealed in our theoretical analysis. 

Five benchmark multi-label datasets\footnote{\url{http://mulan.sourceforge.net/datasets-mlc.html}} are used in our experiments. We choose them because they denote different domains: (i) A music dataset \textbf{CAL500}, (ii) an email dataset \textbf{enron}, (iii) a clinical text dataset \textbf{medical}, (iv) an image dataset \textbf{corel5k}, (v) a tagging dataset \textbf{bibtex}. We randomly split each dataset into two parts, i.e., 70\% for training and 30\% for testing. The experiments are repeated ten times, and the averaged results are reported.

Because our algorithm optimizes a linear model, three linear methods called Binary Relevance (BR) \cite{zhang2014review}, ML-kNN \cite{zhang2007ml} and GFM \cite{WaegemanDJCH14} are provided for fair comparison. As in experiments on synthetic data, we degenerate LIMO ($\lambda_1=\lambda_2=1$) to LIMO-inst ($\lambda_1=0, \lambda_2=1$) and LIMO-label ($\lambda_1=1, \lambda_2=0$). The step size of SGD is set to 0.01. For BR, L2-regularized SVM \cite{CC01a} with C=1 is used as base learner. For ML-kNN and GFM, the number of nearest neighbors is 10. Suitable thresholds discussed in Section \ref{ss:measures_h} are used for classification measures. We take the default parameter settings recommended by authors of the compared methods respectively. Because on one hand, we believe the parameter settings recommended by their authors are meaningful, on the other hand, it is hard to say which parameter setting is better in terms of eleven performance measures.

\begin{figure}[htb] 
 \centering
 \includegraphics[width=8cm]{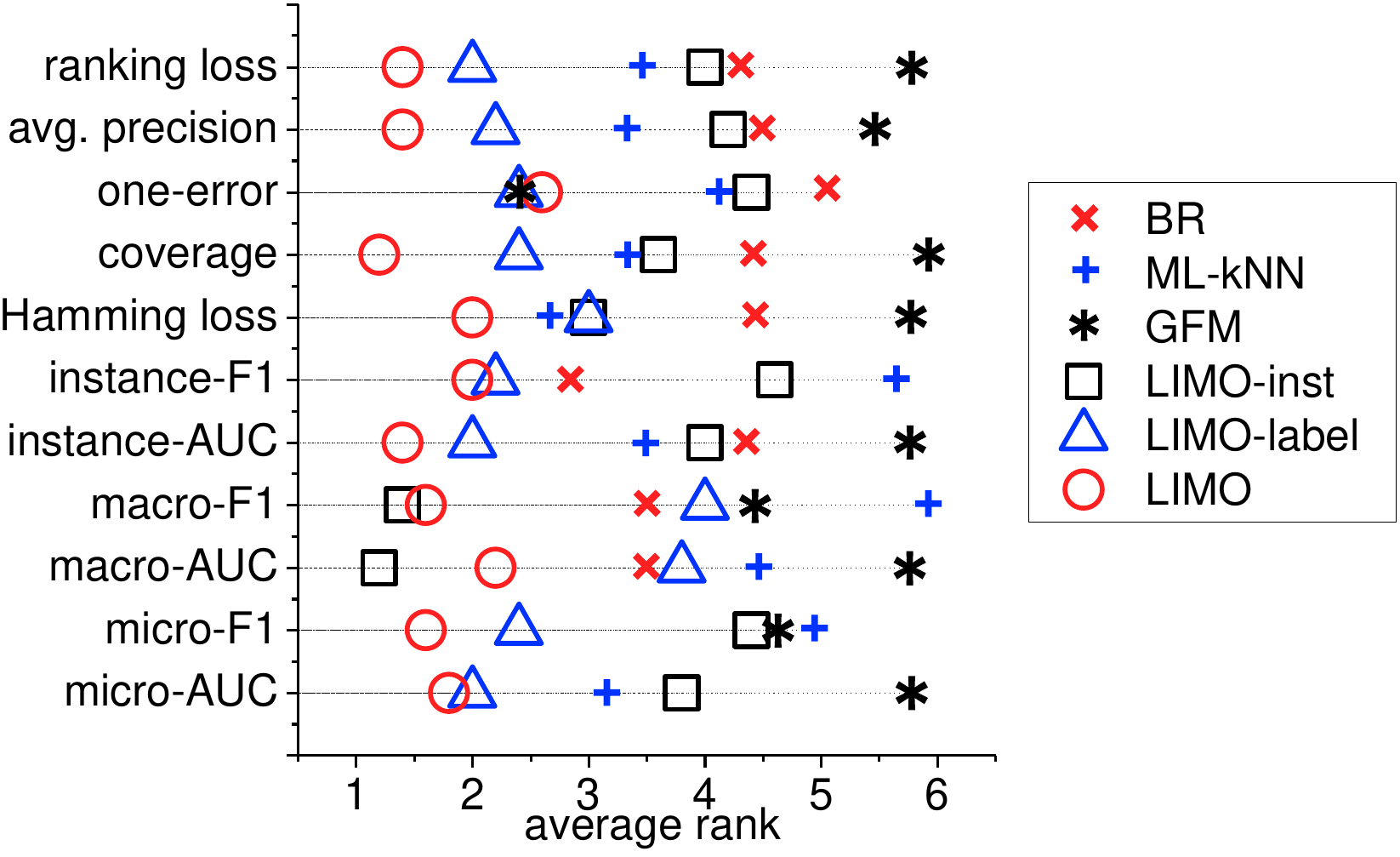} 
 \vspace{-0.4cm}
 \caption{Average rank on benchmark data. The smaller the rank value, the better the performance.} \label{fig:all_measures}
 \vspace{-0.1cm}
\end{figure}

Because some measures are better when higher, and some measures are better when lower, to demonstrate the results more clearly, we compute the average rank of each approach over all datasets on a specific measure. For example, when we want to examine how LIMO performs on ranking loss, we first compute the ranks on each dataset: LIMO ranks 1st on CAL500, enron, bibtex and ranks 2nd on medical, corel5k. Then the average rank of LIMO on ranking loss is (1+1+1+2+2)/5=1.4. Figure \ref{fig:all_measures} shows the average ranks. Due to the space limit, the detailed results used to compute the ranks are provided in Appendix B.2.



The results in Figure \ref{fig:all_measures} are consistent with our theoretical findings. LIMO-inst (the square) performs well on marco-F1 and macro-AUC, while LIMO-label (the triangle) performs well on other performance measures. LIMO (the circle) almost ranks top on every performance measure.

The experiments on synthetic and benchmark data support our theoretical analysis. Although different performance measures focus on different aspects, they share the common property which is formalized in our work as label-wise margin and instance-wise margin. In practice, it is recommended to use higher weight ($\lambda_1/\lambda_2$) on specific margin to optimize the required performance measure. LIMO with nonlinear predictors may perform better, which needs a novel optimization algorithm.

\vspace{-0.15cm}
\section{Conclusion} \label{sec:conclusion}
\vspace{-0.1cm}
In this paper, we establish a unified view for a variety of multi-label performance measures. Based on the proposed concepts of label-wise/instance-wise margins, we prove that some performance measures are to be optimized by label-wise effective classifiers, whereas some by instance-wise effective classifiers. Inspired by the theoretical findings, we design the LIMO approach which can be adjusted to label-wise/instance-wise effective via different parameter settings.  

Our work discloses that there are some shared properties among different subsets of multi-label performance measures. This explains why some measures seem to be redundant in experiments, and suggests that in future empirical studies, rather than randomly grasp a set of measures for evaluation, it is more informative to evaluate using measures with different properties, such as some measures optimized by label-wise effective predictors and some optimized by instance-wise effective predictors. In the future, it is encouraging to study the asymptotic properties of these performance measures when the two margins are suboptimal. The margin view also sheds a light for the design of novel multi-label algorithms.

\section*{Acknowledgements}

This research was supported by the NSFC (61333014), 973 Program (2014CB340501), and the Collaborative Innovation Center of Novel Software Technology and Industrialization. Authors want to thank reviewers for helpful comments, and thank Sheng-Jun Huang, Xiu-Shen Wei, Miao Xu for reading a draft. 


\bibliography{icml2017}
\bibliographystyle{icml2017}
\end{document}


Supplementary material for \textbf{A Unified View of Multi-Label Performance Measures}.

\section{Appendix A: Proofs}
In this section, we provide detailed proofs of the theoretical results presented in the manuscript.
\subsection{Proof of Theorem 4}
Because $F$ is label-wise effective, its order of prediction value on a specific instance $x_i$ is correct. Therefore, the threshold error $\epsilon_i$ can happen in either the two ways:
\begin{enumerate}
\item $\epsilon_i$ positive labels are predicted as negative labels.\\
In this case, the true positive number $TP_i$ on this instance becomes $|Y_{i \cdot}^+|-\epsilon_i$, and the false positive number $FP_i$ is zero, and the false negative number $FN_i$ becomes $\epsilon_i$.\\
The precision value and the recall value will be:
\begin{align*} Prec_i&=\frac{TP_i}{TP_i+FP_i}=1,  \quad Rec_i=\frac{TP_i}{TP_i+FN_i}=\frac{|Y_{i \cdot}^+|-\epsilon_i}{|Y_{i \cdot}^+|}
\end{align*}
And the $F$-measure$_i$ is:
\begin{equation*}
    F\text{-measure}_i=\frac{2Prec_i \times Rec_i}{Prec_i+Rec_i}=\frac{2(|Y_{i \cdot}^+|-\epsilon_i)}{2|Y_{i \cdot}^+|-\epsilon_i}
\end{equation*}
\item $\epsilon_i$ negative labels are predicted as positive labels.\\
In this case, the true positive number $TP_i$ on this instance is still $|Y_{i \cdot}^+|$, and the false positive number $FP_i=\epsilon_i$ , and the false negative number $FN_i$ is zero.\\
The precision value and the recall value will be:
\begin{align*} Prec_i&=\frac{TP_i}{TP_i+FP_i}=\frac{|Y_{i \cdot}^+|}{|Y_{i \cdot}^+|+\epsilon_i},\quad Rec_i=\frac{TP_i}{TP_i+FN_i}=1
\end{align*}
And the $F$-measure$_i$ is:
\begin{equation*}
    F\text{-measure}_i=\frac{2Prec_i \times Rec_i}{Prec_i+Rec_i}=\frac{2|Y_{i \cdot}^+|}{2|Y_{i \cdot}^+|+\epsilon_i}
\end{equation*}
\end{enumerate}
The instance-F1 is lower bounded by the sum of minimum value of $F$-measure$_i$, thus:
\begin{equation*}
  \text{\textit{instance-F1}}(H) \geq \frac{1}{m} \sum_{i=1}^m \min \Big\{\frac{2(|Y_{i \cdot}^+|-\epsilon_i)}{2|Y_{i \cdot}^+|-\epsilon_i}, \frac{2|Y_{i \cdot}^+|}{2|Y_{i \cdot}^+|+\epsilon_i}\Big\}
\end{equation*}
Under the assumption that all the instances are i.i.d drawn, micro-F1 equals instance-F1. Theorem 4 is proved. \qedsymbol
\subsection{Proof of Theorem 5}
Because $F$ is instance-wise effective, its order of prediction value on a specific label $\Y_{\cdot j}$ is correct. Therefore, the threshold error $\epsilon_j$ can happen in either the two ways:
\begin{enumerate}
\item $\epsilon_j$ positive instances are predicted as negative instances.\\
In this case, the true positive number $TP_j$ on this label becomes $|Y_{\cdot j}^+|-\epsilon_j$, and the false positive number $FP_j$ is zero, and the false negative number $FN_j$ becomes $\epsilon_j$.\\
The precision value and the recall value will be:
\begin{align*} Prec_j&=\frac{TP_j}{TP_j+FP_j}=1,  \quad Rec_j=\frac{TP_j}{TP_j+FN_j}=\frac{|Y_{\cdot j}^+|-\epsilon_j}{|Y_{\cdot j}^+|}
\end{align*}
And the $F$-measure$_j$ is:
\begin{equation*}
    F\text{-measure}_j=\frac{2Prec_j \times Rec_j}{Prec_j+Rec_j}=\frac{2(|Y_{j \cdot}^+|-\epsilon_j)}{2|Y_{j \cdot}^+|-\epsilon_j}
\end{equation*}
\item $\epsilon_j$ negative instanaces are predicted as positive instances.\\
In this case, the true positive number $TP_j$ on this label is still $|Y_{\cdot j}^+|$, and the false positive number $FP_j=\epsilon_j$ , and the false negative number $FN_j$ is zero.\\
The precision value and the recall value will be:
\begin{align*} Prec_j&=\frac{TP_j}{TP_j+FP_j}=\frac{|Y_{\cdot j}^+|}{|Y_{ \cdot j}^+|+\epsilon_j},\quad Rec_j=\frac{TP_j}{TP_j+FN_j}=1
\end{align*}
And the $F$-measure$_j$ is:
\begin{equation*}
    F\text{-measure}_j=\frac{2Prec_j \times Rec_j}{Prec_j+Rec_j}=\frac{2|Y_{\cdot j}^+|}{2|Y_{ \cdot j}^+|+\epsilon_j}
\end{equation*}
\end{enumerate}
The macro-F1 is lower bounded by the sum of minimum value of $F$-measure$_j$, thus:
\begin{equation*}
  \text{\textit{macro-F1}}(H) \geq \frac{1}{l} \sum_{j=1}^l \min \Big\{\frac{2(|Y_{\cdot j}^+|-\epsilon_j)}{2|Y_{\cdot j}^+|-\epsilon_j}, \frac{2|Y_{\cdot j}^+|}{2|Y_{\cdot j}^+|+\epsilon_j}\Big\}
\end{equation*}
Theorem 5 is proved. \qedsymbol
\subsection{Proof of LIMO Algorithm}
\begin{mythm} 
In each iteration (step 5 to step 15) of Algorithm 1, the updated direction of the model is an unbiased estimation of the gradient of this objective function:
\begin{equation} \label{obj}
 \begin{split}
     \argmin_{\W, \xi} &\ \sum_{i=1}^l ||\w_i||^2+\lambda_1 \sum_{i=1}^m \sum_{(u,v)}\xi_i^{uv}+\lambda_2 \sum_{j=1}^l \sum_{(a,b)} \xi_{ab}^j \\
     \text{s.t.}\ & \w_u^\top \x_i -\w_v^\top \x_i >1-\xi_i^{uv}, \ \ \xi_i^{uv}\geq 0, \ \text{for }i=1,\cdots,m \text{ and } (u,v)\in Y_{i \cdot}^+ \times Y_{i \cdot}^- \ , \\
                &\w_j^\top \x_a -\w_j^\top \x_b>1-\xi^j_{ab}, \  \ \xi^j_{ab}\geq 0, \ \text{for } j=1,\cdots,l \text{ and } (a,b)\in Y_{\cdot j}^+ \times Y_{\cdot j}^- \ .
\end{split}    
\end{equation} 
\end{mythm}
\begin{spacelessprf}
Suppose that the function in Equation (\ref{obj}) is $f(\W)$, because $\W$  can be decomposed into $[\w_1,\w_2,\cdots,\w_l]$, we consider the partial gradient of a particular $\w_k$:
\begin{equation}
\begin{split}
  \frac{\partial f(\W)}{\partial \w_k}=&2\w_k +\lambda_1 \phi_1 +\lambda_2 \phi_2 = 2\w_k\\
  &+ \lambda_1 \sum_{i=1}^m \Big\{
  \llbracket k\in Y_{i \cdot}^- \rrbracket \x_i  \sum_{j\in Y_{i \cdot}^+ }\llbracket 1-(\w_j - \w_k)^\top \x_i >0 \rrbracket \\
  &\qquad \quad \ \ -\llbracket k\in Y_{i \cdot}^+ \rrbracket \x_i 
   \sum_{j\in Y_{i \cdot}^- }\llbracket 1-(\w_k - \w_j)^\top \x_i >0 \rrbracket \Big\} \\
   &+ \lambda_2 \sum_{a\in Y_{\cdot k}^+} \sum_{ b\in Y_{\cdot k}^-} (\x_b-\x_a) \llbracket 1-\w_k^\top(\x_a-\x_b) >0\rrbracket
  \end{split}
\end{equation}

The second term $\lambda_1 \phi_1$ is the gradient of label-wise margin on $\w_k$, and the third term $\lambda_2 \phi_2$ is the gradient of the instance-wise margin on $\w_k$.

Assume that $(\x_i, y_{ik}, y_{ij})$ is picked in step 5 and 6, the direction will be computed in step 8 or 9 according to:
\begin{equation*}
\begin{split}
  g^{label}(\x_i,y_{ik}, y_{ij})
  =& \llbracket k\in Y_{i \cdot}^- \rrbracket  \lambda_1 \x_i \llbracket 1-(w_k-w_j)^\top \x_i >0\rrbracket \\
  &-\llbracket k\in Y_{i \cdot}^+ \rrbracket  \lambda_1 \x_i \llbracket 1-(w_j-w_k)^\top \x_i >0\rrbracket + \w_k
\end{split}
\end{equation*}
Then do the expectation:
\begin{equation*}
\begin{split}
  E_{\x_{i}}\big[ E_{y_{ij}}[g^{label}(\x_i,y_{ik}, y_{ij})]\big]
  &=\frac{1}{C} E_{\x_{i}}\bigg[ {\lambda_1 \x_i}\sum_{j\in Y_{i \cdot}^+ } \llbracket 1-(w_k-w_j)^\top \x_i >0\rrbracket\\
  &\qquad \quad - \lambda_1 \x_i \sum_{j\in Y_{i \cdot}^- }\llbracket 1-(w_j-w_k)^\top \x_i >0\rrbracket + \frac{1}{D} \w_k\bigg] \\
  &=\frac{1}{C^\prime} \lambda_1 \phi_1  + \frac{1}{D^\prime} \w_k
  \end{split}
\end{equation*}
Where $C^\prime$ and $D^\prime$ are constants. Similarly, we can prove the expectation of the direction in step 11 to 15:
\begin{equation*}
  E_{\x_a,\x_b}[g^{inst}(y_k,\x_a,\x_b)]=\frac{1}{C^{\prime \prime}}\lambda_2 \phi_2 + \frac{1}{D^{\prime\prime}} \w_k
\end{equation*}

Because of the linearity of expectation, and absorbing the constants into $\lambda_1$ and $\lambda_2$, the gradient $\frac{\partial f(\W)}{\partial \w_k}$ can be unbiased estimated. Namely, the updated direction of the algorithm is an unbiased estimation of the gradient of Equation (\ref{obj}).
\end{spacelessprf}
\newpage
\section{Appendix B: Detailed Experimental Results}
In this section, detailed experimental results are included. The results of synthetic data are in Section \ref{ss:syn} and The results of benchmark data are in Section \ref{ss:ben}
\subsection{Detailed Experimental Results of Synthetic Data} \label{ss:syn}
In this section, the detailed experimental results of synthetic data are given.
\begin{table}[!htb]
\centering
\caption{Original absolute value and rescaled value of experiments on ranking measures. In the left columns are absolute values, and in the right columns are rescaled relative values.}
\begin{tabular}{c|c c| c c |c c}
\hline
measure & \multicolumn{2}{c|}{LIMO-inst} & \multicolumn{2}{c|}{LIMO} & \multicolumn{2}{c}{LIMO}\\
\hline
ranking loss & 0.027 & 0.00 & 0.015 & 0.99 & 0.015 & 1.00\\
avg. precision & 0.992 & 0.00 & 0.992 & 0.58 & 0.992 & 1.00\\
one-error & 0.000 & 1.00 & 0.001 & 0.28 & 0.001 & 0.00\\
coverage & 1.576 & 0.00 & 1.557 & 0.97 & 1.556 & 1.00\\
macro-AUC & 0.842 & 1.00 & 0.828 & 0.00 & 0.842 & 0.98\\
instance-AUC & 0.973 & 0.00 & 0.985 & 0.99 & 0.985 & 1.00\\
micro-AUC & 0.861 & 0.14 & 0.854 & 0.00 & 0.903 & 1.00\\
\hline
\end{tabular}
\end{table}

\begin{table}[!htb]
\centering
\setlength{\tabcolsep}{3pt}
\caption{Original absolute value and rescaled value of experiments on classification measures. In the left columns are absolute values, and in the right columns are rescaled relative values.}
\begin{tabular}{c|c c| c c |c c |c c |c c |c c}
\hline
measure & \multicolumn{2}{c|}{LIMO-inst-t} & \multicolumn{2}{c|}{LIMO-inst-t(x)} & 
\multicolumn{2}{c|}{LIMO-label-t} & \multicolumn{2}{c|}{LIMO-label-t(x)} &
\multicolumn{2}{c|}{LIMO-t} & \multicolumn{2}{c}{LIMO-t(x)} \\
\hline
Hamming loss & 0.172 & 0.28 & 0.160 & 0.48 & 0.188 & 0.00 & 0.131 & 1.00 & 0.163 & 0.43 & 0.134 & 0.94\\
micro-F1 & 0.837 & 0.00 & 0.860 & 0.43 & 0.840 & 0.06 & 0.890 & 1.00 & 0.858 & 0.40 & 0.885 & 0.92\\
macro-F1 & 0.869 & 0.87 & 0.857 & 0.25 & 0.861 & 0.46 & 0.859 & 0.35 & 0.872 & 1.00 & 0.852 & 0.00\\
instance-F1 & 0.804 & 0.00 & 0.883 & 0.79 & 0.835 & 0.32 & 0.904 & 1.00 & 0.858 & 0.54 & 0.900 & 0.96\\
\hline
\end{tabular}
\end{table}

\subsection{Detailed Experimental Results of Benchmark Data} \label{ss:ben}
The ranking results in Figure 4 in paper are computed from Table \ref{table:all-detail}. Because this table is too large, we can only rotate it to show in the next page.

\begin{sidewaystable}[!htb]
\centering
\setlength{\tabcolsep}{4pt}
\scriptsize
\caption{Experimental results on eleven multi-label performance measures. For each performance measure, ``$\downarrow$'' indicates ``the smaller the better'' and ``$\uparrow$'' indicates ``the larger the better''. The results are shown in mean$\pm$std(rank). The smaller the rank, the better the performance.
}\label{table:all-detail}
\begin{tabular}{c|c|c c c c c c c c c c c}
\hline
Dataset & Algorithm & hamming loss$\downarrow$& ranking loss$\downarrow$& avg. precision$\uparrow$ & one-error$\downarrow$&  coverage$\downarrow$& instance-F1$\uparrow$ & instance-AUC$\uparrow$ & macro-F1$\uparrow$ & macro-AUC$\uparrow$&micro-F1$\uparrow$ &micro-AUC $\uparrow$\\
\hline
\multirow{6}{*}{CAL500} &BR &.145$\pm$.003(5) &.216$\pm$.005(4) &.470$\pm$.008(4) &.212$\pm$.025(5) &143.025$\pm$2.319(4) &.354$\pm$.009(5) &.784$\pm$.005(4) &.097$\pm$.006(5) &.544$\pm$.012(2) &.357$\pm$.011(5) &.779$\pm$.006(4) \\
&ML-kNN &.139$\pm$.003(3) &.184$\pm$.005(3) &.491$\pm$.007(3) &.106$\pm$.023(3) &129.789$\pm$2.426(2) &.321$\pm$.010(6) &.816$\pm$.005(3) &.053$\pm$.002(6) &.523$\pm$.009(3) &.318$\pm$.010(6) &.813$\pm$.004(3) \\
&GFM &.200$\pm$.002(6) &.522$\pm$.007(5) &.337$\pm$.004(5) &.000$\pm$.000(1) &166.481$\pm$0.906(6) &.454$\pm$.006(3) &.662$\pm$.006(5) &.183$\pm$.005(3) &.518$\pm$.013(5) &.457$\pm$.006(3) &.661$\pm$.006(5) \\
&LIMO-inst &.143$\pm$.004(4) &.545$\pm$.015(6) &.147$\pm$.004(6) &.971$\pm$.022(6) &162.652$\pm$1.539(5) &.386$\pm$.010(4) &.455$\pm$.015(6) &.302$\pm$.008(1) &.566$\pm$.011(1) &.389$\pm$.010(4) &.458$\pm$.014(6) \\
&LIMO-label &.138$\pm$.002(2) &.180$\pm$.004(2) &.499$\pm$.008(2) &.105$\pm$.023(2) &129.993$\pm$2.491(3) &.473$\pm$.004(2) &.820$\pm$.004(2) &.126$\pm$.003(4) &.510$\pm$.011(6) &.477$\pm$.004(2) &.815$\pm$.004(2) \\
&LIMO &.137$\pm$.025(1) &.178$\pm$.004(1) &.501$\pm$.008(1) &.122$\pm$.035(4) &129.323$\pm$2.672(1) &.475$\pm$.006(1) &.822$\pm$.004(1) &.288$\pm$.006(2) &.523$\pm$.011(4) &.479$\pm$.006(1) &.816$\pm$.004(1) \\
\hline
\multirow{6}{*}{medical} &BR &.011$\pm$.001(1) &.073$\pm$.041(5) &.416$\pm$.100(6) &.804$\pm$.029(6) &3.697$\pm$1.752(5) &.766$\pm$.022(1) &.927$\pm$.040(5) &.384$\pm$.040(3) &.877$\pm$.038(3) &.792$\pm$.020(1) &.910$\pm$.039(5) \\
&ML-kNN &.016$\pm$.001(5) &.048$\pm$.008(4) &.788$\pm$.017(4) &.266$\pm$.025(5) &3.034$\pm$0.411(4) &.564$\pm$.033(5) &.953$\pm$.007(4) &.190$\pm$.015(6) &.797$\pm$.029(5) &.654$\pm$.028(4) &.949$\pm$.008(4) \\
&GFM &.025$\pm$.002(6) &.287$\pm$.026(6) &.692$\pm$.025(5) &.217$\pm$.025(3) &6.581$\pm$0.714(6) &.636$\pm$.025(4) &.882$\pm$.013(6) &.216$\pm$.018(4) &.650$\pm$.027(6) &.605$\pm$.027(5) &.875$\pm$.014(6) \\
&LIMO-inst &.015$\pm$.001(4) &.017$\pm$.005(1) &.881$\pm$.018(2) &.170$\pm$.028(2) &1.248$\pm$0.279(1) &.444$\pm$.090(6) &.983$\pm$.005(1) &.448$\pm$.024(2) &.901$\pm$.032(1) &.439$\pm$.087(6) &.979$\pm$.005(1) \\
&LIMO-label &.014$\pm$.001(3) &.032$\pm$.006(3) &.829$\pm$.018(3) &.217$\pm$.026(4) &2.237$\pm$0.378(3) &.641$\pm$.030(3) &.968$\pm$.006(3) &.207$\pm$.012(5) &.859$\pm$.035(4) &.702$\pm$.020(3) &.960$\pm$.007(3) \\
&LIMO &.013$\pm$.001(2) &.019$\pm$.006(2) &.893$\pm$.017(1) &.147$\pm$.027(1) &1.423$\pm$0.350(2) &.706$\pm$.019(2) &.981$\pm$.006(2) &.464$\pm$.024(1) &.896$\pm$.029(2) &.757$\pm$.012(2) &.977$\pm$.006(2) \\
\hline
\multirow{6}{*}{enron} &BR &.070$\pm$.003(6) &.136$\pm$.010(4) &.539$\pm$.086(4) &.533$\pm$.263(6) &16.834$\pm$0.669(4) &.482$\pm$.009(3) &.866$\pm$.010(4) &.187$\pm$.015(3) &.631$\pm$.025(5) &.473$\pm$.008(3) &.814$\pm$.008(4) \\
&ML-kNN &.053$\pm$.001(3) &.096$\pm$.004(3) &.624$\pm$.014(3) &.310$\pm$.022(4) &13.615$\pm$0.423(3) &.409$\pm$.022(5) &.904$\pm$.004(3) &.083$\pm$.008(6) &.633$\pm$.022(4) &.461$\pm$.017(4) &.898$\pm$.003(1) \\
&GFM &.069$\pm$.003(5) &.554$\pm$.027(6) &.399$\pm$.021(6) &.246$\pm$.041(2) &31.645$\pm$0.764(6) &.428$\pm$.022(4) &.669$\pm$.014(6) &.118$\pm$.012(5) &.553$\pm$.015(6) &.437$\pm$.016(5) &.654$\pm$.011(6) \\
&LIMO-inst &.054$\pm$.001(4) &.205$\pm$.008(5) &.520$\pm$.008(5) &.344$\pm$.013(5) &23.679$\pm$0.804(5) &.404$\pm$.053(6) &.796$\pm$.008(5) &.310$\pm$.018(1) &.717$\pm$.015(1) &.414$\pm$.056(6) &.810$\pm$.006(5) \\
&LIMO-label &.049$\pm$.001(2) &.085$\pm$.003(2) &.672$\pm$.010(2) &.233$\pm$.017(1) &12.324$\pm$0.444(2) &.565$\pm$.011(1) &.916$\pm$.003(2) &.137$\pm$.005(4) &.644$\pm$.019(3) &.591$\pm$.008(2) &.897$\pm$.003(2) \\
&LIMO &.049$\pm$.001(1) &.083$\pm$.003(1) &.672$\pm$.010(1) &.253$\pm$.022(3) &11.880$\pm$0.255(1) &.562$\pm$.010(2) &.918$\pm$.003(1) &.278$\pm$.017(2) &.663$\pm$.021(2) &.596$\pm$.006(1) &.896$\pm$.004(3) \\
\hline
\multirow{6}{*}{corel5k} &BR &.014$\pm$.000(5) &.280$\pm$.010(5) &.077$\pm$.013(6) &.962$\pm$.004(6) &207.643$\pm$3.477(5) &.139$\pm$.005(4) &.720$\pm$.010(5) &.044$\pm$.003(4) &.605$\pm$.004(4) &.158$\pm$.006(3) &.706$\pm$.009(5) \\
&ML-kNN &.009$\pm$.000(1) &.135$\pm$.002(3) &.245$\pm$.004(2) &.736$\pm$.009(3) &114.727$\pm$1.658(3) &.017$\pm$.002(6) &.865$\pm$.002(3) &.009$\pm$.001(6) &.540$\pm$.007(5) &.027$\pm$.003(6) &.866$\pm$.002(3) \\
&GFM &.021$\pm$.001(6) &.803$\pm$.012(6) &.100$\pm$.005(5) &.516$\pm$.030(1) &320.449$\pm$2.173(6) &.150$\pm$.008(3) &.416$\pm$.012(6) &.029$\pm$.002(5) &.516$\pm$.006(6) &.146$\pm$.011(4) &.411$\pm$.013(6) \\
&LIMO-inst &.010$\pm$.000(2) &.275$\pm$.004(4) &.105$\pm$.004(4) &.897$\pm$.006(5) &172.120$\pm$2.183(4) &.058$\pm$.003(5) &.725$\pm$.004(4) &.118$\pm$.003(1) &.706$\pm$.006(1) &.057$\pm$.002(5) &.725$\pm$.004(4) \\
&LIMO-label &.011$\pm$.000(4) &.112$\pm$.003(1) &.289$\pm$.006(1) &.710$\pm$.010(2) &99.629$\pm$2.136(2) &.214$\pm$.004(1) &.888$\pm$.003(1) &.050$\pm$.002(3) &.658$\pm$.006(3) &.236$\pm$.004(1) &.882$\pm$.002(1) \\
&LIMO &.011$\pm$.000(3) &.116$\pm$.003(2) &.227$\pm$.005(3) &.791$\pm$.008(4) &94.253$\pm$2.109(1) &.152$\pm$.005(2) &.884$\pm$.003(2) &.117$\pm$.004(2) &.692$\pm$.007(2) &.177$\pm$.005(2) &.881$\pm$.003(2) \\
\hline
\multirow{6}{*}{bibtex} &BR &.016$\pm$.001(5) &.114$\pm$.007(3) &.528$\pm$.012(2) &.428$\pm$.014(2) &32.758$\pm$1.929(4) &.399$\pm$.009(1) &.886$\pm$.007(3) &.318$\pm$.016(3) &.866$\pm$.007(4) &.419$\pm$.012(3) &.869$\pm$.007(4) \\
&ML-kNN &.014$\pm$.000(2) &.218$\pm$.004(5) &.339$\pm$.006(5) &.599$\pm$.007(6) &56.259$\pm$1.260(5) &.160$\pm$.007(6) &.782$\pm$.004(5) &.066$\pm$.006(6) &.661$\pm$.007(5) &.211$\pm$.007(5) &.776$\pm$.005(5) \\
&GFM &.037$\pm$.000(6) &.707$\pm$.003(6) &.210$\pm$.004(6) &.492$\pm$.010(5) &85.281$\pm$0.701(6) &.223$\pm$.008(5) &.618$\pm$.003(6) &.130$\pm$.007(5) &.575$\pm$.003(6) &.185$\pm$.007(6) &.626$\pm$.003(6) \\
&LIMO-inst &.014$\pm$.000(1) &.120$\pm$.003(4) &.494$\pm$.008(4) &.469$\pm$.016(4) &32.403$\pm$0.640(3) &.392$\pm$.005(2) &.880$\pm$.003(4) &.323$\pm$.005(2) &.921$\pm$.002(2) &.438$\pm$.006(1) &.877$\pm$.003(3) \\
&LIMO-label &.014$\pm$.000(4) &.071$\pm$.002(2) &.527$\pm$.008(3) &.433$\pm$.013(3) &20.425$\pm$0.422(2) &.386$\pm$.007(4) &.929$\pm$.002(2) &.232$\pm$.004(4) &.911$\pm$.002(3) &.405$\pm$.007(4) &.917$\pm$.002(2) \\
&LIMO &.014$\pm$.000(3) &.058$\pm$.001(1) &.570$\pm$.004(1) &.390$\pm$.008(1) &17.447$\pm$0.357(1) &.390$\pm$.007(3) &.942$\pm$.001(1) &.326$\pm$.006(1) &.924$\pm$.002(1) &.435$\pm$.004(2) &.938$\pm$.002(1) \\
\hline

\end{tabular}
\end{sidewaystable}